\documentclass[journal]{IEEEtran}
\usepackage{multirow}
\usepackage{graphicx}
\usepackage{booktabs}
\usepackage{amsmath}
\usepackage[numbers]{natbib}
\usepackage[T1]{fontenc}
\usepackage{xcolor, soul}

\ifCLASSINFOpdf
\else
\fi

\newcommand{\RN}[1]{%
  \textup{\uppercase\expandafter{\romannumeral#1}}%
}
\begin{document}
%
\title{A Probabilistic Framework for Estimating the Risk of Pedestrian-Vehicle Conflicts at Intersections}
%
%
%

\author{Pei Li,
        Huizhong Guo,
        Shan Bao,~\IEEEmembership{Member,~IEEE}~and~Arpan Kusari,~\IEEEmembership{Member,~IEEE}
\thanks{P. Li, H.Guo, S. Bao and A. Kusari are with University of Michigan Transportation Research Institute, University of Michigan, Ann Arbor,
MI, 48109 USA e-mail: {peili@knights.ucf.edu, hzhguo, shanbao, kusari}@umich.edu.}
\thanks{Manuscript received July 27, 2022}}

%
%

\markboth{Journal of \LaTeX\ Class Files}%
{Shell \MakeLowercase{\textit{et al.}}: Bare Demo of IEEEtran.cls for IEEE Journals}
%



\maketitle

\begin{abstract}
Pedestrian safety has become an important research topic among various studies due to the increased number of pedestrian-involved crashes. To evaluate pedestrian safety proactively, surrogate safety measures (SSMs) have been widely used in traffic conflict-based studies  as they do not require historical crashes as inputs. However, most existing SSMs were developed based on the assumption that road users would maintain constant velocity and direction. Risk estimations based on this assumption are less unstable, more likely to be exaggerated, and unable to capture the evasive maneuvers of drivers. Considering the limitations among existing SSMs, this study proposes a probabilistic framework for estimating the risk of pedestrian-vehicle conflicts at intersections. The proposed framework loosen restrictions of constant speed by predicting trajectories using a Gaussian Process Regression and accounts for the different possible driver maneuvers with a Random Forest model. Real-world LiDAR data collected at an intersection was used to evaluate the performance of the proposed framework. The newly developed framework is able to identify all pedestrian-vehicle conflicts. Compared to the Time-to-Collision, the proposed framework provides a more stable risk estimation and captures the evasive maneuvers of vehicles. Moreover, the proposed framework does not require expensive computation resources, which makes it an ideal choice for real-time proactive pedestrian safety solutions at intersections. 
\end{abstract}


%
\IEEEpeerreviewmaketitle

\section{Introduction}
%
%
%
%

\IEEEPARstart{P}{edestrian} safety is an important consideration in the transportation system design, due to the vulnerability of this group when involved in traffic crashes. The pedestrian fatality rate has increased by about $46\%$ from 2011 to 2020, and pedestrian-involved crashes have become more deadly and more frequent \cite{pointtraffic}. There were 6,516 pedestrian fatalities that occurred on the roadways in the United States in 2020, which represents a $3.9\%$ increase from the number in 2019. Moreover, $15 \%$ of the pedestrian fatalities occurred at intersections~\cite{pointtraffic}.

Over the past decades, different studies have been conducted to investigate pedestrian safety at intersections, which can be divided into two categories based on the data sources used in the studies: crash-based and conflict-based studies. Crash-based studies usually use historical crashes to predict future pedestrian-related crashes, analyze crash severity, etc. \cite{preusser2002pedestrian, lee2005comprehensive, abdel2005exploring, pulugurtha2011pedestrian, schneider2010association}. Differently, conflict-based studies use driving data to develop behavior matrices to predict or identify pedestrian-vehicle conflicts and quantify the risk of conflicts \cite{ismail2009automated, ismail2010automated, wu2018novel, zhang2020prediction}. The benefit of using conflicts over actual crashes is they can proactively evaluate the status of traffic safety and quantify the risks with detailed information from both vehicles and pedestrians. Moreover, since pedestrian-vehicle crashes are relatively rare compared to vehicle-vehicle crashes, it may take much more time to collect a sufficient amount of pedestrian-involved crashes with necessary pre-crash information. Lastly, conflicts between pedestrians and vehicles may not actually cause crashes, but still, need to be analyzed to better understand interactions between different road users. 

Various SSMs have been used in conflict-based studies, including time to collision (TTC) \cite{hayward1972near}, post encroachment time (PET) \cite{cooper1984experience}, gap time \cite{archer2004methods}, deceleration to safety time (DST) \cite{hyden1987development}, etc. Recently, due to the development of object detection techniques, it is much easier to get trajectories of road users at intersections using devices like cameras~\cite{zhang2020prediction, zaki2013application, ismail2010automated}, LiDARs~\cite{wu2018novel, ma2022virtual}, etc. Using trajectories obtained by these devices, various SSMs can be estimated and used to estimate the risk of pedestrian-vehicle conflicts at intersections. However, most SSMs are developed upon the assumption of constant velocity and moving direction, which is invalid for vehicles at intersections since they frequently adjust their speed and acceleration to accommodate traffic conditions and their turning maneuvers. Moreover, due to this assumption, metrics like TTC are not able to capture the evasive behaviors of vehicles and overestimate the actual risk of conflicts~\cite{ismail2010automated}. Although SSMs estimated using the actual trajectories (e.g., PET) do not suffer from this assumption, they can only be applied in post-analytical studies and thus cannot predict conflicts in advance. Lastly, a threshold needs to be decided while using SSMs such as TTC and PET, which can be subjective to each individual study.


Considering the aforementioned limitations of existing SSMs, this study proposes a probabilistic framework to estimate the risk of pedestrian-vehicle conflicts at intersections. The proposed framework improves the existing SSMs by first relaxing the assumption of constant velocity and direction. The framework is able to accurately predict the trajectory of a vehicle and estimate the probability of it taking different maneuvers. Second, the estimated risk from the proposed framework can reflect the evasive maneuvers of the vehicle, which cannot be captured by existing SSMs such as TTC. Third, the framework does not require a subjective choice of thresholds, which provides a more generic way of analyzing the relationships between the risk of conflicts and interactions between pedestrians and vehicles. Lastly, the proposed framework uses simple observations (i.e., position and velocity) as inputs and does not require heavy computational resources compared with deep learning methods. This makes it significantly efficient to be deployed in real-time proactive pedestrian safety solutions at intersections.

The rest of the paper is organized as follows: Section \RN{2} reviews the current SSMs and their applications in existing studies. Section \RN{3} introduces the methods used in this paper, including data preparation and the proposed framework. Section \RN{4} presents the results of the proposed framework, as well as a comparison to existing SSMs. The conclusions and future research directions are summarized in Section \RN{5}.

\section{Literature Review}

SSMs can be divided into two categories based on the estimation approach: proximity-based and deceleration-based SSMs. Proximity-based SSMs are more popular among the existing studies. The definitions of some proximity-based SSMs, including TTC, PET, and gap time are shown below:
\begin{itemize}
    \item TTC can be defined as the time that remains until a conflict between two road users occurs given that the conflict course and speed difference are maintained \cite{hayward1972near}.
    \item PET is the time difference between the moment an "offending" road user passes out of the area of a potential conflict and the moment of arrival at the potential conflict point by the "conflicted" road user possessing the right-of-way \cite{cooper1984experience}.
    \item Gap time is a variation of PET, which is calculated at each time instant by projecting the movement of road users that have the potential conflicts~\cite{archer2004methods}.
\end{itemize}

\citet{ismail2009automated} applied TTC, PET, and gap time to analyze pedestrian-vehicle conflicts using video data. Although these SSMs were able to identify most pedestrian-vehicle conflicts, the authors suggested that combining different SSMs could be the choice to better capture pedestrian-vehicle conflicts. Some studies adopted this strategy and used multiple SSMs. For example, \citet{wu2020improved} combined PET with other SSMs (i.e., the proportion of the stopping distance and crash potential index) to identify near-crash events between pedestrians and vehicles.  The main reason for using additional SSMs was because PET could not reflect the change of driver behaviors, such as a sudden deceleration before reaching the potential conflict point.

Deceleration-based SSMs, such as DST, deceleration rate (DR), deceleration rate required to stop (DRS), identify pedestrian-vehicle conflicts based on vehicle deceleration and are defined as:
\begin{itemize}
    \item DST is defined as the deceleration a vehicle needed to reach a non-negative PET value if the movements of the conflicting road users remain unchanged \cite{hyden1987development}.
    \item DR is the highest rate at which a vehicle must decelerate to avoid a conflict \cite{olszewski2020surrogate}.
    \item DRS is the constant deceleration rate required for the vehicle to stop and give the right-of-way to pedestrians \cite{fu2018novel}.
\end{itemize}  
\citet{fu2018novel} used DRS to evaluate pedestrian safety at unsignalized intersections. Results suggested that stop sign-controlled intersections had lower DRS than unprotected intersections, which indicated that stop signs provided better protections for pedestrians. \citet{olszewski2020surrogate} used the value of deceleration to identify pedestrian-vehicle conflicts at intersections. A threshold of $4m/s^2$ was used to detect the abrupt braking behaviors of vehicles. \citet{ismail2010automated} conducted a before-and-after study to evaluate the impact of the pedestrian scramble phasing on pedestrian safety. The number of medium and high values of DST significantly decreased after the implementation of the pedestrian scramble phasing, indicating the positive impact of the scramble phasing.

Existing SSMs have been successfully applied by different studies. Nevertheless, little research has been done to deal with the inherent flaws of existing SSMs. First, SSMs such as TTC suffer from the assumption of constant velocity and direction and overestimate the severity of actual conflicts. For example, a typical conflict happens when the vehicle is making a right turn and the pedestrian is crossing. The vehicle's speed is constantly changing during the process of turning, which may not be captured by the TTC due to this assumption. Second, \citet{johnsson2018search} indicated that TTC and PET may not be able to capture the evasive actions of vehicles (e.g, a vehicle brakes as approaching a pedestrian). Lastly, deceleration-based measures may falsely identify a conflict since they focus solely on the magnitude of evasive actions. Thus, it is necessary to develop a comprehensive and generic framework for estimating the risk of pedestrian-vehicle conflicts at intersections, which can relax the existing assumption and identify the vehicle's evasive behaviors.

\section{Methods}
\subsection{Data Description}
\begin{figure}[!htbp]
    \centering
    \includegraphics[scale=0.6]{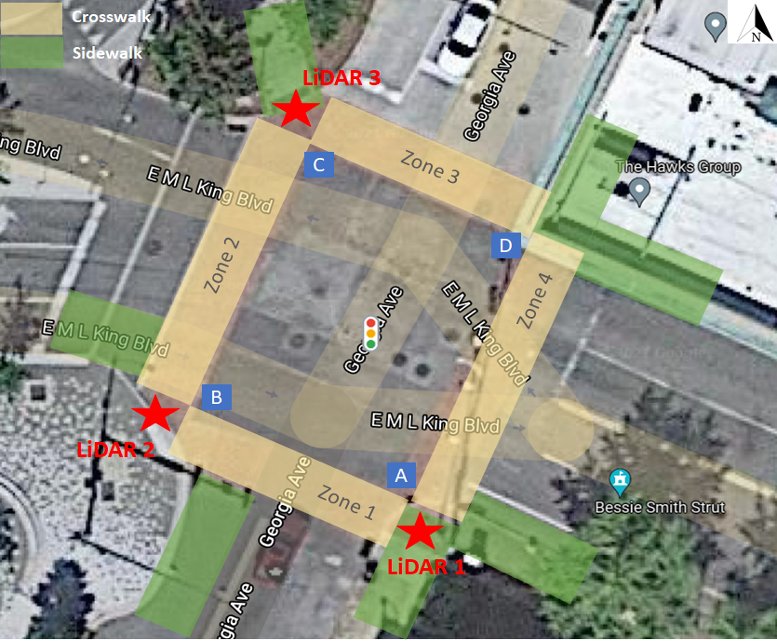}
    \caption{LiDAR set-up at the studied intersection}
    \label{fig:map}
\end{figure}

Data used in this study was collected by three Ouster OS1-128 LiDAR sensors installed at the intersection of E. MLK Blvd. and Georgia Ave. in Chattanooga, Tennessee (Fig.~\ref{fig:map}) from approximately 11 AM to 12 PM on Oct 8, 2021. The data update frequency was 10Hz. The studied intersection is a signalized intersection with designated pedestrian signals and marked crosswalks. The LiDAR data was initially processed by the Seoul Robotics software and the object-level data including labels, corresponding size, velocity, signal information, etc. was released as part of the Transportation Research Board (TRB) Transportation Forecasting Competition 2022~\citep{transfor22}.  

\subsection{Data Preprocessing}
The goal of the data preprocessing is to retain a set of clean and relatively complete vehicle and pedestrian trajectories for the forthcoming analysis. 
We describe the steps taken to identify data issues, clean data anomalies, and transform the data into a readily usable format below:

\subsubsection{Creating unique trajectory label for each individual tracked object} A unique label (i.e., vehicle, pedestrian, cyclists, or misc) was assigned for every trajectory based on a majority vote, where each trajectory was given the label with the highest proportion of assignments. 

\subsubsection{Estimating the start and end points of the four crosswalks}
\begin{figure}
    \centering
    \includegraphics[scale=0.3]{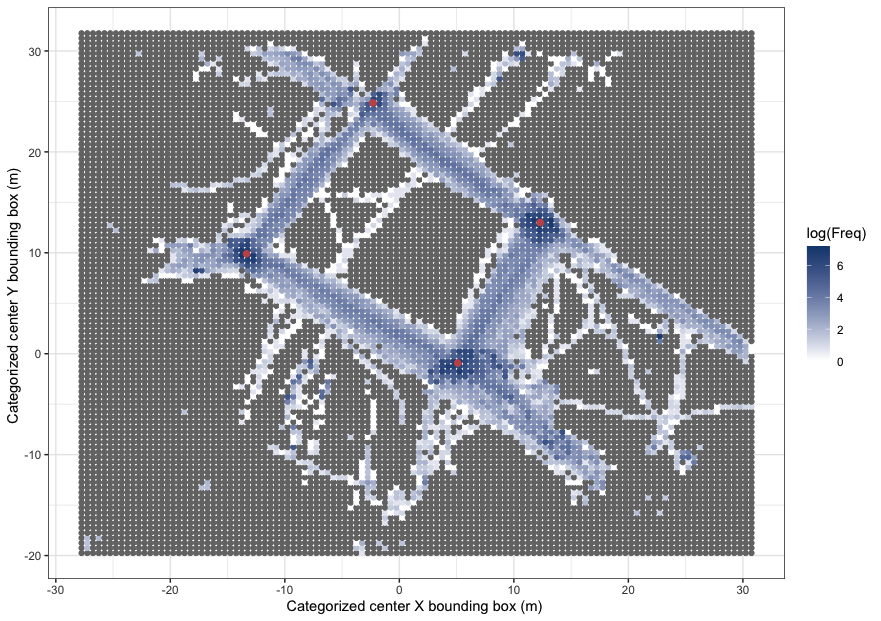}
    \caption{Squares colored by log-transformed pedestrian trajectory density, with squares never visited by any pedestrian trajectories in grey; identified crosswalk start/end points were denoted by red dots}
    \label{fig:xwalk_start_end}
\end{figure}

This study focuses on interactions and conflicts between vehicles and pedestrians at an intersection. The position of crosswalks and the movement of road users relative to the crosswalk is thereby an important component for any forthcoming analysis. Due to the absence of GPS coordinates, we located the start and end of crosswalks based on pedestrian trajectories. Since the majority of pedestrians used the crosswalk to cross the intersection and their movements were captured beyond the crosswalk, pooling all pedestrian trajectories gives us a rough estimate of the areas containing the start/end of the four crosswalks. Within each estimation, the start/end of a crosswalk was the center of the area with the highest frequency of trajectories passing through (Fig. \ref{fig:xwalk_start_end}).



It is recognized that the identified start and end points were not necessarily the center points of the actual crosswalks. Rather, they are the center of areas at the two ends of crosswalks with the most pedestrian trajectories passing by. Thus, the identified points are considered closely approximate the actual crosswalk start/end location while also accounting for the real movement of pedestrians at this intersection.

\subsubsection{Classifying entering and moving direction for vehicle trajectories}
The two pairs of crosswalks' start/end points in the diagonal direction were used to divide the intersection area into four quadrants. 
The entering direction of a specific vehicle trajectory was then determined based on which quadrant the first point of the trajectory belongs to. For example, if the first point locates in the bottom right quadrant, the trajectory was labeled entering from the South of the intersection. Additionally, the sequence of unique quadrants a trajectory has traveled through was extracted and used to label the moving direction for all vehicle trajectories. Especially, this paper only focused on through, left-turn, and right-turn movements since they accounted for most movements.  



\subsubsection{Merging pedestrian trajectories}

Preliminary data analysis showed that pedestrian trajectories were oftentimes short and suspected to be incomplete. Thus, the goal of this step is to identify trajectories that were potentially from the same pedestrian and combine them into a single new trajectory. 

A trajectory $T_B$ is considered an immediately next trajectory of trajectory $T_A$ if $T_A$'s end point and $T_B$'s start point satisfy the following four criteria:
\begin{itemize}
    \item Difference in time is no more than 0.2 secs.
    
    \item Difference in distance (based on X and Y coordinates) is no more than 1 m.
    
    \item Difference in heading is no more than 90\textdegree.
    
    \item Difference in the angle of a trajectory, pointing from the position of the first data point to the last data points, is no more than 120\textdegree.
\end{itemize}

All thresholds in the four criteria were selected based on an examination of the data and were intentionally kept tight to reduce false matching of trajectories. A slightly larger threshold was set for heading to accommodate the wider variation of headings for pedestrian trajectories. 
If multiple candidates satisfy the four criteria, the one that outperformed the others (i.e., with smaller differences in the given criteria) was selected.

\subsubsection{Selecting pedestrian trajectories}
Since a large proportion of short and incomplete trajectories still exist even after merging, additional steps were taken to remove pedestrian trajectories that were:
\begin{itemize}
    \item Too short in time ($\leq$1 sec) or distance ($\leq$ 5 m)
    \item With a higher proportion ($\geq$50\%) of invalid data points
    \item Moving too fast to be considered a pedestrian, i.e., velocity is no less than 3m/s for at least 10 data points (approximately 1 sec)
    \item Either outside the crosswalk or the roadway areas, or were within the roadway area but leaving the crosswalk
\end{itemize}

\subsection{Probabilistic Risk Estimation Framework}

\begin{figure}[thpb]
    \centering
    \includegraphics[scale=.3]{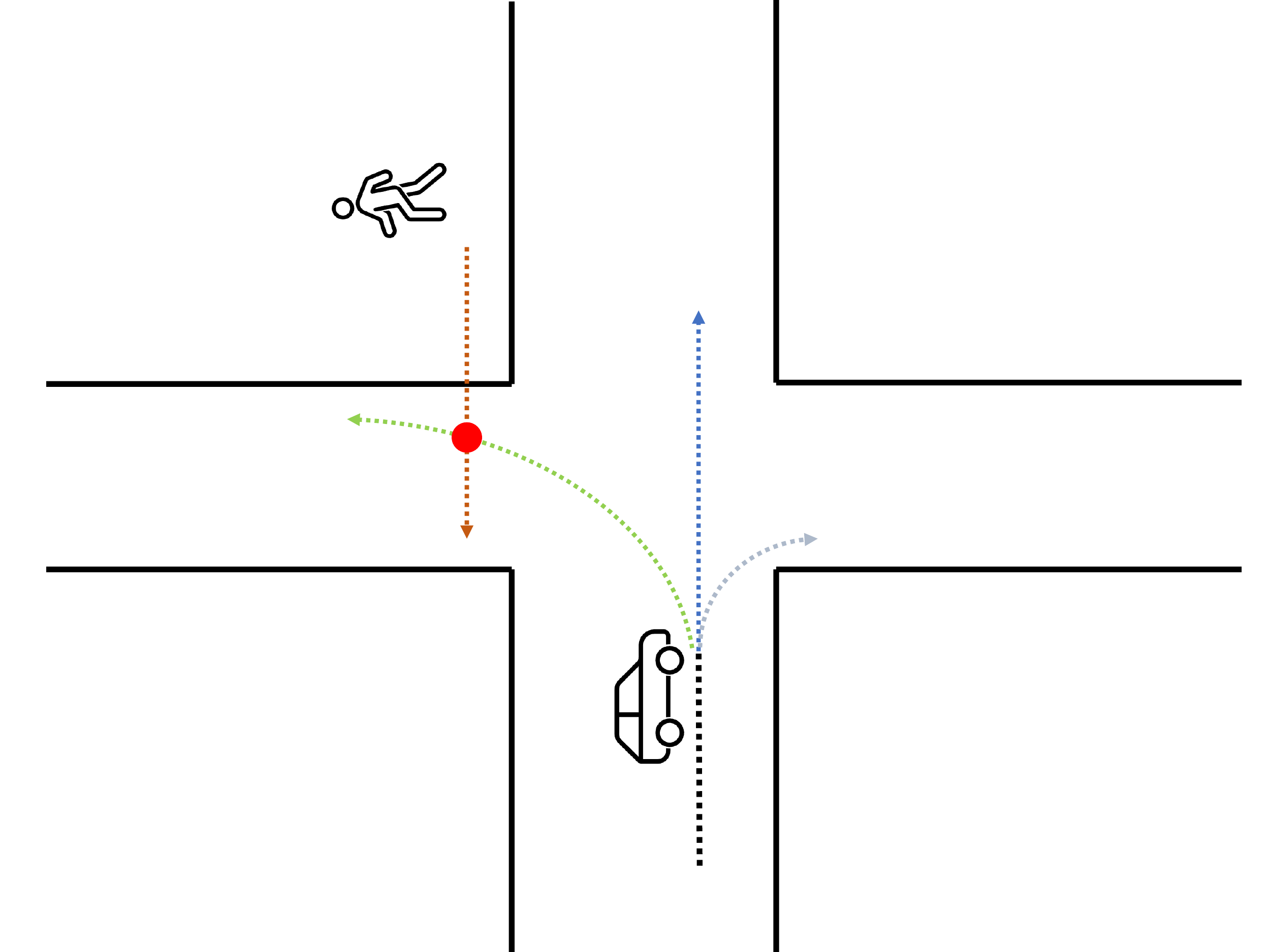}
    \caption{An example of a pedestrian-vehicle conflict}
    \label{fig:risk_illustration}
\end{figure}

Fig.~\ref{fig:risk_illustration} shows an example of a potential conflict between a vehicle and a pedestrian. Given the current position of a vehicle, the proposed framework needs to estimate the probability of the vehicle making different maneuvers (left turn, right turn, and going straight) and the corresponding future trajectories to estimate the risk of this conflict. Moreover, the framework should also be able to predict the pedestrian's future trajectory. Then, given the predicted vehicle and pedestrian trajectories, the risk of the conflict can be estimated using Equation~\ref{eq:3}, where $i=1,2,3$ denotes the vehicle's maneuvers of turning left, turning right, and going straight, respectively, and $Prob(i|pos)$ represents the probability of a vehicle making the $i$th maneuver given its current position ($pos$). The risk of a vehicle-pedestrian conflict given vehicle maneuver $i$ ($Risk_{i}$) is calculated as the exponential of the negative TTC (Equation~\ref{eq:4}) so that a smaller value of TTC corresponds to a higher risk of conflict. $T_{veh_{i}}$ and $T_{ped_{i}}$ represent the time a vehicle and pedestrian needs to reach the conflict point ($c_{i}$) if it exists. The use of exponential function allows the risk to decay gracefully with the increase in time difference without the explicit need for a threshold. 

\begin{equation}\label{eq:3}
\begin{aligned}
   Risk = \sum_{i=1}^{N} Risk_{i} \cdot Prob(i|pos)
\end{aligned}
\end{equation}

\begin{equation}\label{eq:4}
\begin{aligned}
   Risk_{i} = 
   \begin{cases}
    exp(-|T_{veh_{i}} - T_{ped}|),&\text{if}\;c_{i}\;\text{exists}\\
    0,              & \text{otherwise}
    \end{cases}
\end{aligned}
\end{equation}

\begin{figure}[thpb]
    \centering
    \includegraphics[scale=.3]{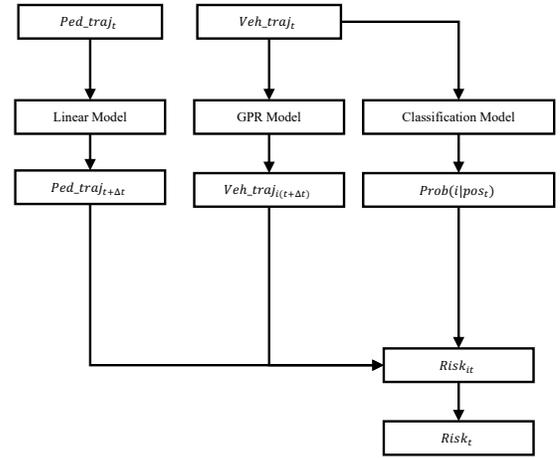}
    \caption{The flowchart of estimating the risk of conflicts}
    \label{fig:flowchart of framework}
\end{figure}

Fig.~\ref{fig:flowchart of framework} shows the flowchart of risk estimation at timestep $t$. The GPR model is applied to the vehicle for predicting its future trajectory in terms of the $i$th maneuver, while the classification model is used to estimate the probability of the vehicle making the $i$th maneuver. A linear dynamic model is applied to the pedestrian trajectory for estimating the future trajectory, which assumes the pedestrian will keep a constant velocity. The reason for using a complicated model for predicting the vehicle's trajectory is because a vehicle usually has more complex behaviors than a pedestrian, such as turning, accelerating, braking, etc, which makes the assumption of the constant velocity not valid. Then the $Risk_{it}$ is estimated for the $i$th vehicle maneuver at time $t$ using Equation~\ref{eq:4}, while the risk at time $t$ ($Risk_t$) is estimated using Equation~\ref{eq:3}.

\subsection{Gaussian Process Regression}
Gaussian process regression (GPR) is a non-parametric Bayesian approach \cite{schulz2018tutorial}, which has been widely used for modeling trajectory data \cite{8500614, aoude2011mobile}. GPR balances functional complexity with capturing the underlying data and is thus both more general and more principled than other forms of regression~\cite{cox2012gaussian}. A GPR model on an input $X$ can be defined by its mean function $m(X)$ and covariance function $k(X, X')$ as shown in Equation~\ref{eq:1}.

\begin{equation}\label{eq:1}
\begin{aligned}
   f \sim gp(m(X), k(X, X'))
\end{aligned}
\end{equation}

\subsubsection{Kernels}
Two kernels (i.e., covariance functions) were commonly used in existing studies, including the Radial Basis Function (RBF) and Rational Quadratic (RQ) kernels as shown in Equations~\ref{eq:2} and~\ref{eq:8}. The performance of these two kernels was compared in this study to select the best kernel.
\begin{equation}\label{eq:2}
\begin{aligned}
   k_{RBF}(X, X') = exp(\frac{(X-X')^2}{2\sigma^2})
\end{aligned}
\end{equation}

\begin{equation}\label{eq:8}
\begin{aligned}
   k_{RQ}(X, X') = (1 + \frac{(X-X')^2}{2\alpha\sigma^2})^{-\alpha}
\end{aligned}
\end{equation}

\subsubsection{Hyperparameters}
The kernel used in a GPR model usually has several hyperparameters that are unknown and need to be inferred from the data \cite{schulz2018tutorial}. For example, the RBF kernel in Equation~\ref{eq:2} has one hyperparameter $\sigma$, which is its length scale. The RQ kernel in Equation~\ref{eq:8} has two hyperparameters including $\alpha$ and $\sigma$, which are the length scale and weighting parameter, respectively.


\subsubsection{Loss function}
Similar to other machine learning models, the development of a GPR model involves training and testing. Moreover, since the kernel of a GPR model usually has several hyperparameters, they should be optimized during the process of training. All of these require a specific loss function to measure the performance of a GPR model. The log marginal likelihood is used to compute the loss of a GPR model. Given a GPR model (Equation~\ref{eq:1}), its inputs $X$, and targets $y$, the marginal likelihood of the model is estimated using Equation~\ref{eq:9} \cite{gardner2018gpytorch}, while the log transformation of the marginal likelihood is negated to be used as the loss of a GPR model. Moreover, an Adam optimizer \cite{kingma2014adam} with a learning rate of 0.1 was used in this study to optimize hyperparameters of a GPR model based on its loss.

\begin{equation}
\label{eq:9}
\begin{aligned}
    \mathcal{L} = p_{f}(y|X) = \int p(y|f(X))p(f(X)|X)df
\end{aligned}
\end{equation}

Given a vehicle's position $pos = (x_{t}, y_{t})$ and velocity $v = (v_{x_{t}}, v_{y_{t}})$ at time $t$. A pair of GPR models are used to model its speed based on its current position as shown in Equation~\ref{eq:5}. 

\begin{equation}
\label{eq:5}
\begin{aligned}
   v_{x_{t}} \sim gp_{x}(m(x_{t}, y_{t}), k((x_{t}, y_{t}),(x_{t}, y_{t})')) \\
   v_{y_{t}} \sim gp_{y}(m(x_{t}, y_{t}), k((x_{t}, y_{t}),(x_{t}, y_{t})')) 
\end{aligned}
\end{equation}

To predict a vehicle's future trajectory based on its current position $(x_{0}, y_{0})$, the pair of GPR models is firstly applied to sample a trajectory derivative (speed) $(v_{x_{0}}, v_{y_{0}})$ at the current moment, the next position $(x_{1}, y_{1})$ is estimated as $(x_{0}+v_{x_{0}} \Delta t, y_{0}+v_{y_{0}} \Delta t)$, while $\Delta t = t_1 - t_0$. Then the next position $(x_{2}, y_{2})$ is estimated using the speed $(v_{x_1}, v_{y_1})$ sampled from $(x_{1}, y_{1})$. This process is repeated for $T$ time steps ($\Delta t$) to predict the vehicle's trajectory from $pos(x_{1}, y_{1})$ to $pos(x_{T}, y_{T})$.

\section{Results}
\subsection{Data Summary}
\begin{figure}[thpb]
    \hspace{-1cm}
    \centering
    \includegraphics[scale=0.4]{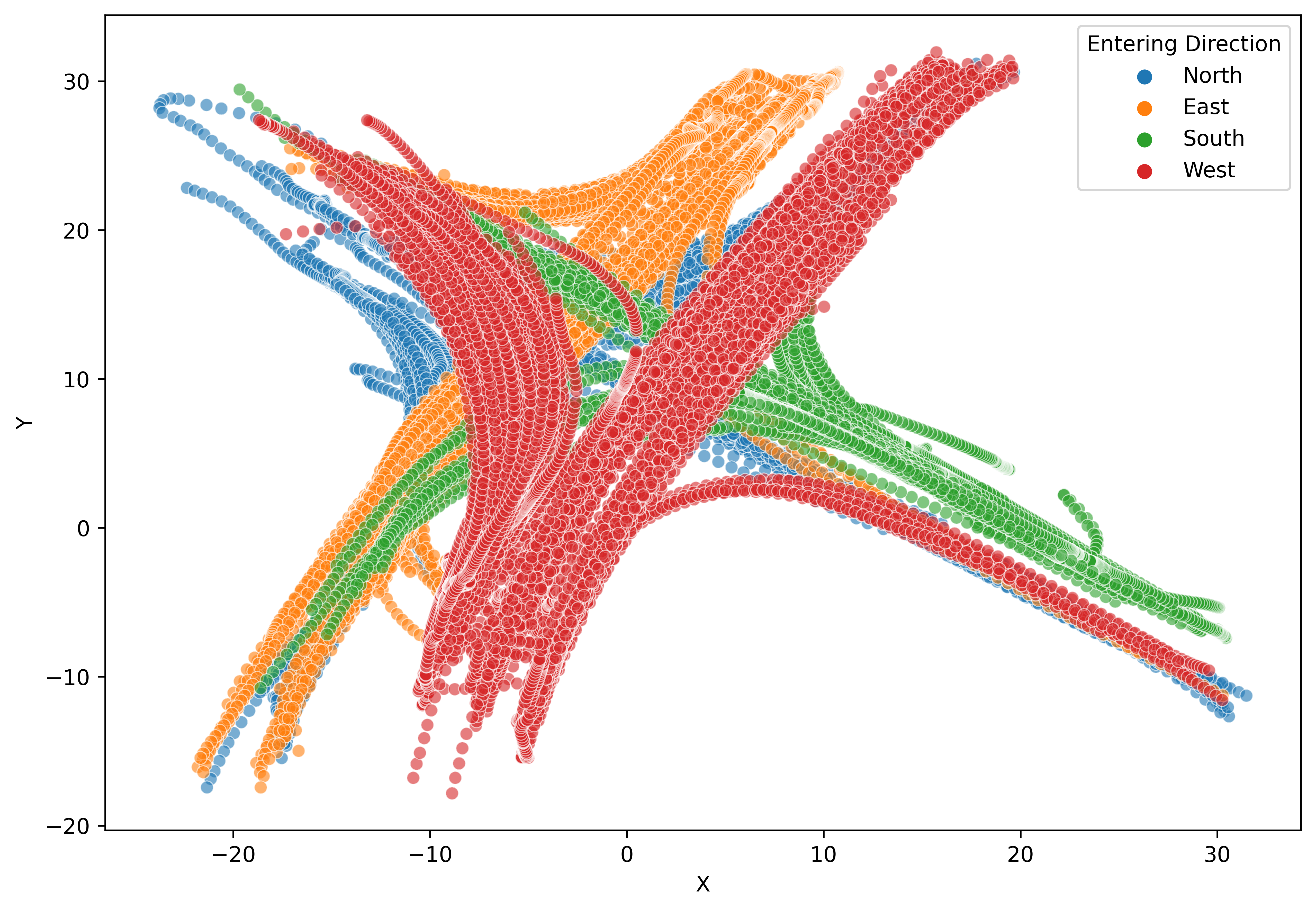}
    \caption{Trajectories of vehicles entering from different directions}
    \label{fig:veh_traj}
\end{figure}

After data preparation, the data set contains trajectories of 935 vehicles and 311 pedestrians. Table~\ref{table:data_descriptive} shows the descriptive statistics of the vehicle and pedestrian data. Moreover, the number of vehicles turning left, turning right, and going straight are 145, 199, and 591, respectively Fig.~\ref{fig:veh_traj} shows trajectories of vehicles entering from different directions. In total, the number of vehicles entering the intersection from the North, East, South, and West directions are 197, 351, 93, and 294, respectively.

\begin{table*}[ht] 
\caption{Descriptive Statistics} 
\centering      
\begin{tabular}{llp{6cm}llll} 
\hline                       
Type & Name & Description & Mean & Std & Min & Max\\ [0.5ex] 
\hline 
Vehicle & $Velocity_x$ & The speed of vehicles in X-axis (m/s) & -0.42 & 3.23 & -13.02 & 15.65\\    
                 & $Velocity_y$ & The speed of vehicles in Y-axis (m/s) & 0.31 & 4.45 & -13.58 & 15.60\\
                 & $Yaw rate$ & The yaw rate of vehicles (deg/s) & 3.05 & 1.56 & 0.00 & 6.28 \\
\addlinespace
Pedestrian    & $Velocity_x$ & The speed of pedestrians in X-axis (m/s) & 0.03 & 0.79 & -6.38 & 6.63 \\    
                 & $Velocity_y$ & The speed of pedestrians in Y-axis (m/s) & -0.02 & 0.82 & -6.21 & 6.56\\
\hline     
\end{tabular} 
\label{table:data_descriptive}  
\end{table*}

\subsection{Framework Development}
\subsubsection{Vehicle maneuvers identification}
\begin{figure}[thpb]
    \centering
    \includegraphics[scale=.4]{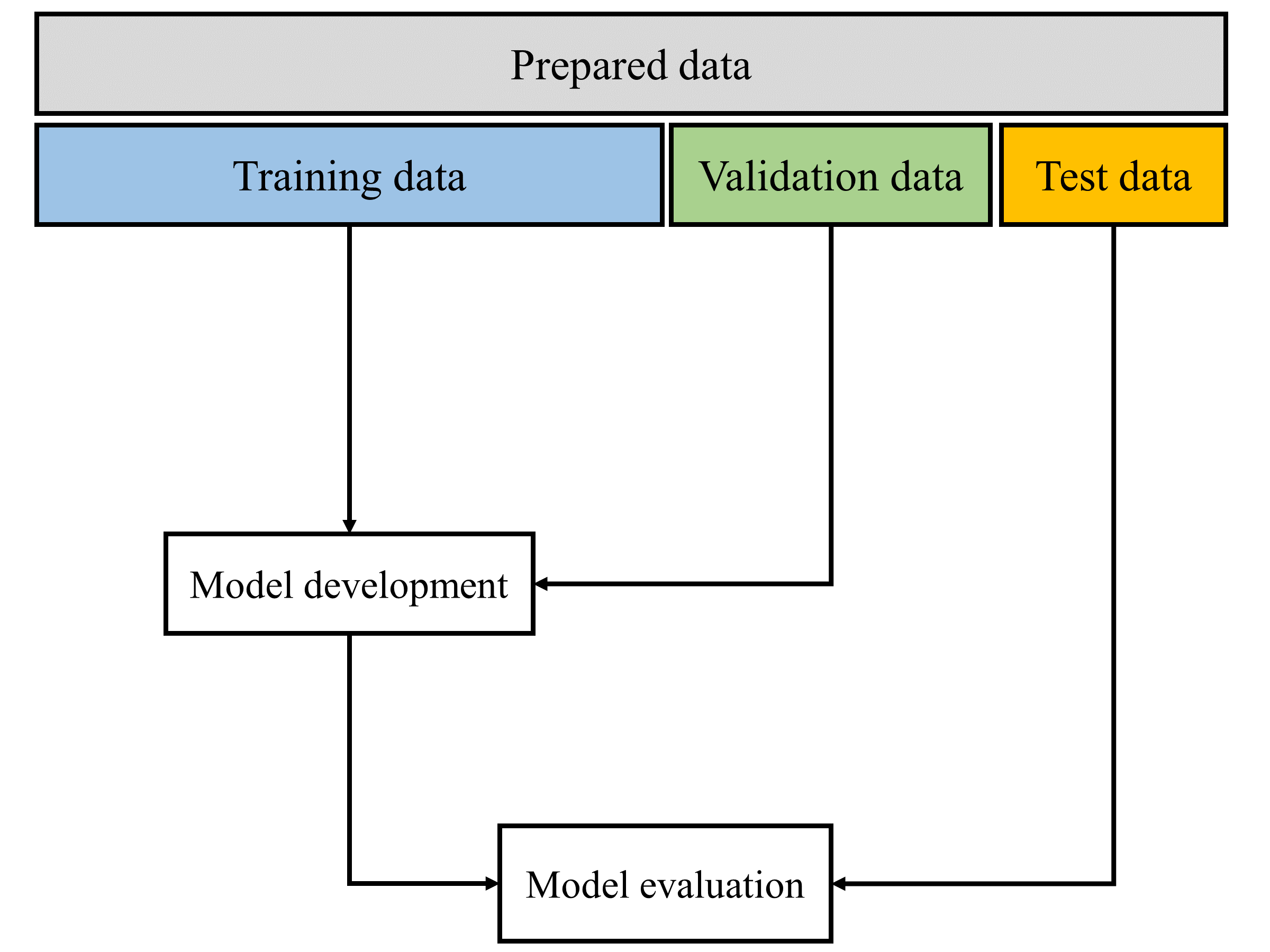}
    \caption{The process of model development}
    \label{fig:model_development}
\end{figure}

Random forest was used to predict the probability of the vehicle making different maneuvers. Five features were used as the model's inputs, including the vehicle's location, velocity, yaw rate, and entering direction. The prepared trajectory data were randomly divided into training, validation, and test in a ratio of 80\%:10\%:10\% for 10 times. The process of developing the model is shown in Fig.~\ref{fig:model_development}. The model was developed based on the training and validation data by tuning its hyperparameters, then it was evaluated on the test data. Moreover, since the data contained more going straight than turning maneuvers, directly developing the model on the data could make it favor the majority maneuver. An over-sampling method was applied to balance the training data. Synthetic Minority Oversampling Technique (SMOTE)~\cite{chawla2002smote} was used in this paper since it was widely used for a similar purpose and achieved better results compared with other over-sampling methods. 

The model's results on the test data are shown in Table~\ref{table:precision_recall_rf}. Precision, recall, and F1-Score were averaged over the 10 splits. Equations of estimating these metrics can be found in \cite{scikit-learn}.

\begin{table}[ht] 
\caption{Accuracy metrics for identifying maneuvers} 
\centering      
\begin{tabular}{l c c c}  
\hline                       
Maneuvers & Precision & Recall & F1-score\\ 
\hline                    
Left-turn  & 0.938 & 0.961 & 0.950 \\    
Right-turn & 0.916 & 0.946 & 0.931  \\
Going straight & 0.973 & 0.957 & 0.965  \\ 
\hline     
\end{tabular} 
\label{table:precision_recall_rf}  
\end{table}

\subsubsection{Vehicle trajectory prediction}
\begin{figure*}[thpb]
    \hspace{-1cm}
    \centering
    \includegraphics[scale=0.35]{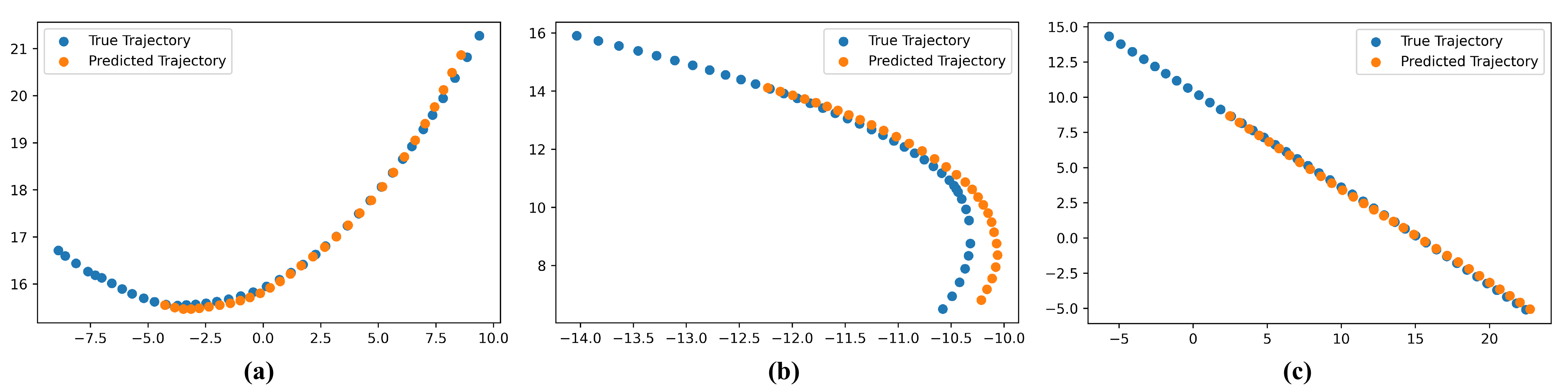}
    \caption{Examples of GPR models on predicting trajectory, (a) vehicle is turning left, (b) vehicle is turning right, (c) vehicle is going straight}
    \label{fig:gp_results}
\end{figure*}

Trajectories of vehicles were first divided into 12 clusters based on their turning maneuvers and entering directions. A pair of GPR models were developed using Equation~\ref{eq:5} for each cluster. Then the developed GPR models were applied to predict vehicle trajectory. Moreover, the RQ kernel was used in this study as it achieved better results than the RBF kernel. Fig.~\ref{fig:gp_results} shows three examples of using the developed GPR models for predicting vehicle trajectory. Specifically, the developed GPR models were used to predict the next 30 trajectory points using only the information of the 10th trajectory point. Results from Fig.~\ref{fig:gp_results} indicated that GPR models were able to accurately predict future vehicle trajectory using data from only one point. Moreover, GPR models had a better performance in the earlier stage of predictions, this was because the accumulation of errors could lead to worse results in the later stage of predictions.

A dynamic model was used as the baseline to compare with the GPR model. The dynamic model uses a vehicle's current speed and acceleration to estimate its next position, as shown in Equation~\ref{eq:6}, where $a_{x_t}$ and $a_{y_t}$ represents the acceleration of a vehicle at time $t$ in X- and Y-axis, $v_{x_t}$ and $v_{y_t}$ represents the velocity of a vehicle at time $t$ in X- and Y-axis, $(x_t, y_t)$ represents the position of a vehicle at time $t$. 

\begin{equation}
\label{eq:6}
\begin{aligned}
   x_{t+\Delta t} = \frac{1}{2}a_{x_t}{\Delta t}^2 + v_{x_t}\Delta t + x_{t} \\
   y_{t+\Delta t} = \frac{1}{2}a_{y_t}{\Delta t}^2 + v_{y_t}\Delta t + y_{t}
\end{aligned}
\end{equation}

To quantify the performance of a model on predicting vehicle trajectory, Euclidean distance between the estimated trajectory point $(\hat{x_{i}}, \hat{y_{i}})$ and the real trajectory point $(x_{i}, y_{i})$ was used to quantify the error of trajectory prediction as shown in Equation~\ref{eq:7}. Table~\ref{table:gpr_dynamic_start_points} shows the results of GPR and dynamic models on predicting the next 30 points (three seconds) using different starting points in terms of different maneuvers. To estimate the results in Table~\ref{table:gpr_dynamic_start_points}, when the starting point was the 10th point, GPR models were applied to estimate the next 30 trajectory points for each vehicle, then the distance between the estimated and true trajectory points was estimated using Equation~\ref{eq:7}. The mean and standard deviation of the distance of all vehicles making the same maneuver were then estimated. The same process was used to get the results of dynamic models. Table~\ref{table:gpr_dynamic_start_points} indicates that the GPR model has achieved much better results than the dynamic model. First, the GPR model had higher accuracy than the dynamic model as it had smaller values of mean distance. Second, the GPR model had a much more stable performance compared with the dynamic model since it had much smaller values of the standard deviation of distance. Lastly, the GPR model was not sensitive to the choice of starting points since it remained a consistent performance among different starting points.

\begin{equation}
\label{eq:7}
\begin{aligned}
   Distance = \sqrt{(\hat{x_{i}}-x_{i})^2+(\hat{y_{i}}-y_{i})^2}
\end{aligned}
\end{equation}

Moreover, to evaluate the impact of the prediction horizon on a model's performance, the 10th trajectory point of a vehicle was used as the input for the GPR and dynamic model to predict the next 10, 15, and 20 points. Table~\ref{table:gpr_dynamic_end_points} shows the results of GPR and dynamic models on different prediction horizons. The performance of both models was highly related to the choice of prediction horizons, while the GPR model achieved a more stable performance than the dynamic model. For example, the mean distance of the GPR model ranged from 0.6m to 1.3m, but the mean distance of the dynamic model increased from 2.4m to 4.6m. Results from Table~\ref{table:gpr_dynamic_end_points} suggested that the GPR model was more robust towards the change of prediction horizons as it had a much smaller value of the standard deviation of distance, which made it more suitable for predicting vehicle trajectory than a dynamic model.

\begin{table*}[ht]
\caption{Results of GPR and Dynamic Models on Different Starting Points} 
\centering
\begin{tabular}{llllll}
\hline
\multirow{2}{*}{Starting Points} & \multirow{2}{*}{Maneuvers} & \multicolumn{2}{l}{GPR Model Results} & \multicolumn{2}{l}{Dynamic Model Results}\\
\cline{3-6}
& & Mean & Std & Mean & Std\\
\hline
\multirow{3}{*}{10} & Left-turn & 1.307 & 1.611 & 6.741 & 6.316 \\
                    & Right-turn & 2.169 & 2.076 & 7.863 & 6.443 \\
                    & Going Straight & 2.035 & 2.022 & 6.890 & 5.326 \\
\hline
\multirow{3}{*}{15} & Left-turn & 1.869 & 2.069 & 6.087 & 5.051 \\
                    & Right-turn & 2.201 & 2.151 & 7.120 & 6.347 \\
                    & Going Straight & 2.141 & 2.196 & 6.056 & 4.921 \\
\hline
\multirow{3}{*}{20} & Left-turn & 1.679 & 2.024 & 5.863 & 4.927 \\
                    & Right-turn & 2.020 & 2.008 & 6.247 & 5.915 \\
                    & Going Straight & 2.526 & 2.375 & 4.933 & 4.344 \\
\hline
\label{table:gpr_dynamic_start_points}
\end{tabular}
\end{table*}

\begin{table*}[ht]
\caption{Results of GPR and Dynamic Models on Different Prediction Horizons} 
\centering
\begin{tabular}{llllll}
\hline
\multirow{2}{*}{Prediction Horizons} & \multirow{2}{*}{Maneuvers} & \multicolumn{2}{l}{GPR Model Results} & \multicolumn{2}{l}{Dynamic Model Results}\\
\cline{3-6}
& & Mean & Std & Mean & Std\\
\hline
\multirow{3}{*}{10} & Left-turn & 0.635 & 0.645 & 2.384 & 2.261 \\
                    & Right-turn & 0.676 & 0.582 & 2.765 & 2.212  \\
                    & Going Straight & 0.702 & 0.539 & 2.424 & 1.860  \\
\hline
\multirow{3}{*}{15} & Left-turn & 0.803 & 0.898 & 3.490 & 3.344 \\
                    & Right-turn & 1.008 & 0.959 & 4.066 & 3.297 \\
                    & Going Straight & 1.021 & 0.848 & 3.566 & 2.751 \\
\hline
\multirow{3}{*}{20} & Left-turn & 0.941 & 1.137 & 4.589 & 4.381 \\
                    & Right-turn & 1.387 & 1.367 & 5.348 & 4.366 \\
                    & Going Straight & 1.332 & 1.184 & 4.685 & 3.622 \\
\hline
\label{table:gpr_dynamic_end_points}
\end{tabular}
\end{table*}

\subsection{Risk Estimation}

To evaluate the performance of the proposed framework in estimating the risk of pedestrian-vehicle conflict, real pedestrian-vehicle conflicts were first identified from the data using PET. PET was used as it was based on actual trajectories and was suggested as the most reliable SSM~\cite{ismail2009automated}. A threshold of 3 seconds was used in this paper for PET. In total, 16 pedestrian-vehicle conflicts were identified from the data. The proposed framework was then applied to the data. A potential conflict was identified if the value of an estimated risk was larger than 0. Several metrics were used to evaluate the performance of the framework, including sensitivity, false alarm rate, and the area under the curve (AUC). The estimation of these metrics can be referred to~\cite{li2020deep}. The proposed framework had a sensitivity of 1, a false alarm rate of 0.19, and an AUC of 0.93. Results indicated that the proposed framework was able to predict all conflicts while maintaining a relatively low false alarm rate. 

\begin{figure}[thpb]
    \hspace{-1cm}
    \centering
    \includegraphics[scale=0.45]{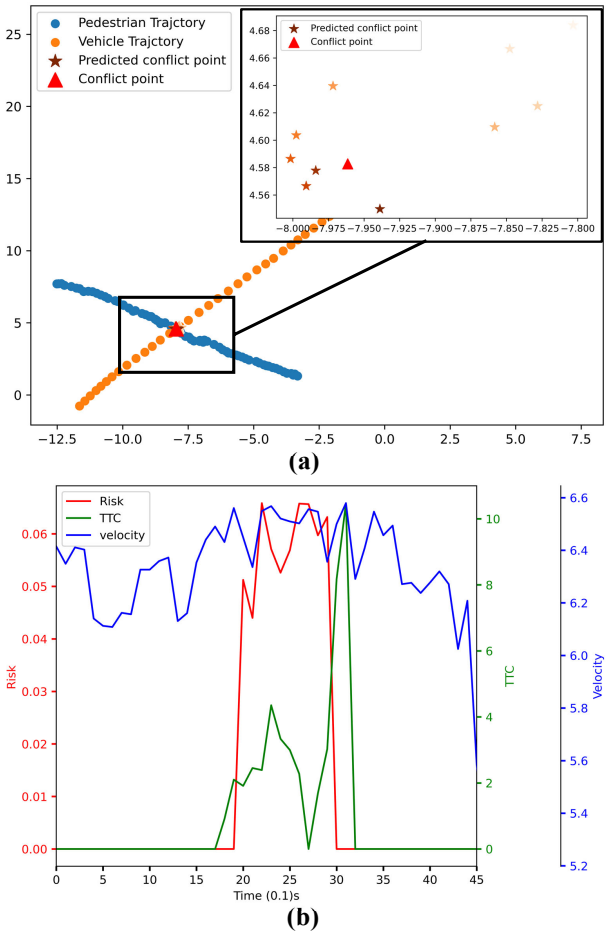}
    \caption{Results of the 1st case study. (a) Trajectories of the pedestrian and vehicle, as well as the predicted and actual conflict points, The change of colors from light to dark represents the increase of time, (b) The change of the risk, TTC, and vehicle's velocity over time.}
    \label{fig:case_study_1}
\end{figure}

Moreover, two case studies were conducted to compare the performance of the proposed risk framework and the traditional TTC. Fig.~\ref{fig:case_study_1} shows one case study of a pedestrian-vehicle conflict while the vehicle was going straight. Specifically, Fig.~\ref{fig:case_study_1} (a) shows the trajectory of the pedestrian and vehicle, the actual conflict point, and predicted conflict points using the proposed framework. Fig.~\ref{fig:case_study_1} (b) shows the risk estimated using the proposed framework, TTC, and vehicle's velocity. Velocity was plotted since it is a crucial factor in terms of estimating the risk of a pedestrian-vehicle conflict. Since the vehicle was going straight, its trajectory was relatively easy to predict. Therefore, the predicted conflict points overlapped with the actual conflict point.
Fig.~\ref{fig:case_study_1} (a) also shows the distribution of actual and predicted conflict points in a zoomed-in view. The predicted conflicts were getting closer to the actual conflict point, which corresponded to the results from Table~\ref{table:gpr_dynamic_end_points}. In this case, the vehicle was going straight and the assumption of constant velocity and direction may not be violated, therefore, both the proposed risk and TTC were able to predict the conflict as evidenced by Fig.~\ref{fig:case_study_1} (b), although TTC was much less stable than the proposed risk. 

\begin{figure}[thpb]
    \hspace{-1cm}
    \centering
    \includegraphics[scale=0.45]{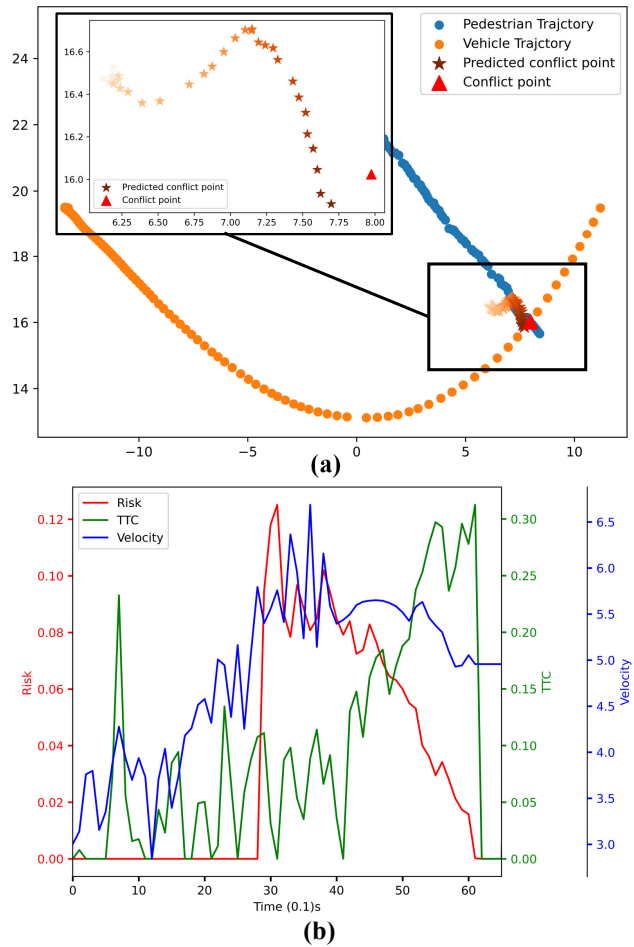}
    \caption{Results of the 2nd case study. (a) Trajectories of the pedestrian and vehicle, as well as the predicted and actual conflict points, The change of colors from light to dark represents the increase of time, (b) The change of the risk, TTC, and vehicle's velocity over time.}
    \label{fig:case_study_2}
\end{figure}

Fig.~\ref{fig:case_study_2} shows another example in which the vehicle was making a left turn and had a potential conflict with the pedestrian. Fig.~\ref{fig:case_study_2} (a) shows locations of the predicted and actual conflict points. A wider variation among the predicted conflict points could be observed as compared with Fig.~\ref{fig:case_study_1} (a). However, the zoomed-in view shows that the predicted conflict points were getting closer to the actual conflict point with the increase in time. This observation indicated that the performance of the proposed risk framework gradually improved as the vehicle moving. Moreover, since the vehicle was making a turning maneuver, the assumption of constant velocity and direction was violated. Therefore, the values of TTC were extremely unstable and had significant fluctuations as shown in Fig.~\ref{fig:case_study_2} (b). The values of the proposed risk were much more stable than the traditional TTC due to the ability of the GPR model to accurately predict the vehicle's trajectory. Moreover, Fig.~\ref{fig:case_study_2} (b) shows that the proposed risk framework could capture the evasive maneuvers of the vehicle as the value of estimated risk gradually decreased once reached the highest point. This decreasing trend was consistent with the decrease in the vehicle's velocity. As the vehicle was approaching the pedestrian, the vehicle would make certain evasive behaviors and therefore reduce its velocity, which decreased the overall risk of the conflict. However, the value of TTC did not reflect any observable patterns and had much larger fluctuations compared with the estimated values of the risk, which made it harder to analyze interactions between the pedestrian and vehicle using TTC.

In summary, the proposed risk framework outperforms the traditional TTC in several ways. First, the proposed risk framework is able to capture the vehicle's evasive behaviors due to the prediction ability of the GPR model. Second, the proposed risk framework is more robust toward noisy observations and can achieve lower false alarm rates compared with the traditional TTC. Third, the proposed risk framework is able to accurately predict the conflict point between the pedestrian and vehicle. Lastly, the proposed risk indicator is not a discrete metric since it does not require a subjective threshold, which can help researchers better understand the pattern of conflict risk and interactions between pedestrians and vehicles.

\section{Conclusions}

Extensive studies have been conducted to use various SSMs for improving pedestrian safety at intersections over the past decade. However, existing SSMs suffer from the assumption of constant velocity and direction, pre-defined thresholds, and other limitations, which make pedestrian-vehicle conflicts prediction and risk estimation lack of high accuracy.

This paper proposed a probabilistic framework to estimate the risk of pedestrian-vehicle conflicts at intersections. LiDAR sensor data has been used to extract trajectories of pedestrians and vehicles at an intersection. Trajectory data were then used to illustrate the performance of the proposed framework in predicting the vehicle's trajectory and estimating the risk of conflicts. The unique contributions of the proposed framework are: (1) it is able to accurately estimate the real-time risk of pedestrian-vehicle conflicts given their current position; (2) it can capture and filter the evasive behaviors of the vehicle; and (3) it has achieved more stable and interpretative results compared to traditional TTC. In addition, it is able to predict all conflicts with the information presented in the trajectory data, and only simple inputs (i.e. position and velocity) are used to operate and have the potential to expand to new locations.

In the future, additional studies could be conducted to improve the performance of the trajectory prediction algorithm, as well as develop methods for predicting the trajectory of pedestrians. The transferability of the proposed framework should be investigated using data from other intersections.

\section*{Acknowledgment}
The authors would like to thank the Artificial Intelligence and Advanced Computing Applications Committee of the Transportation Research Board (TRB) for providing the data. All results and opinions are those of the authors only and do not reflect the opinion or position of TRB.

\ifCLASSOPTIONcaptionsoff
  \newpage
\fi



%

\bibliographystyle{IEEEtranN}
\bibliography{trb_template}
%

\begin{IEEEbiography}[{\includegraphics[width=1in,height=1.25in,clip,keepaspectratio]{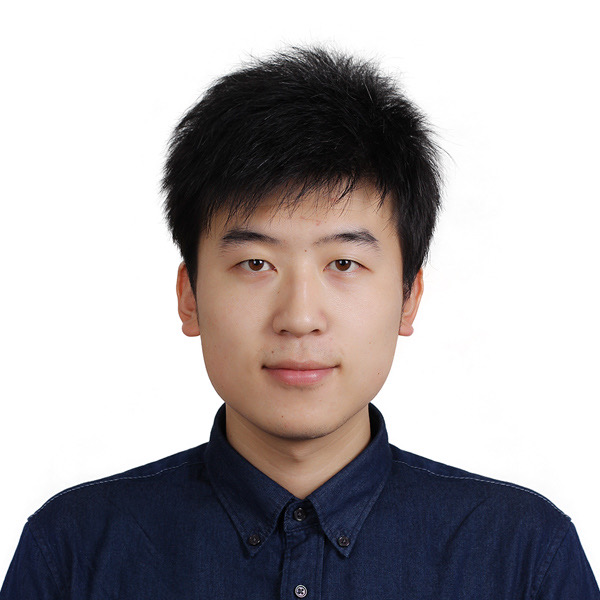}}]{Pei Li} 
is a postdoctoral research fellow at the University of Michigan Transportation Research Institute. He received his Ph.D. in Civil Engineering from the University of Central Florida, Orlando, FL, USA. His research interests include traffic safety, deep learning, connected vehicles, and traffic simulation.
\end{IEEEbiography}

\begin{IEEEbiography}[{\includegraphics[width=1in,height=1.25in,clip,keepaspectratio]{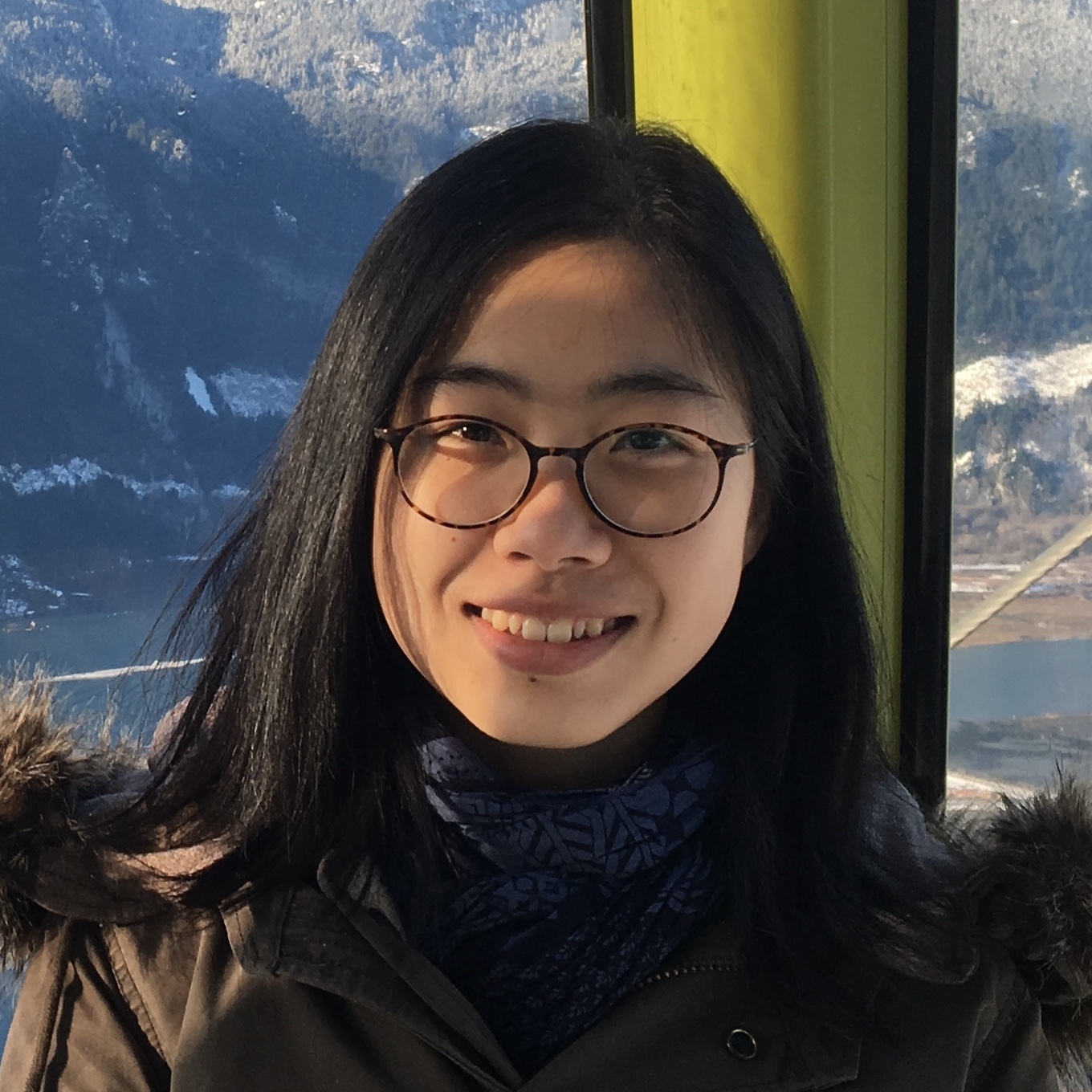}}]{Huizhong Guo}
is a postdoctoral research fellow in the Human Factors Group at the University of Michigan Transportation Research Institute. She received her Ph.D. in Civil Engineering and M.S. in Statistics from the University of Washington, USA. Her research focuses on driver behavior and its implication on transportation safety. She has extensive experience working with naturalistic driving data and laboratory data through statistical and machine learning methods. 
\end{IEEEbiography}


\begin{IEEEbiography}[{\includegraphics[width=1in,height=1.25in,clip,keepaspectratio]{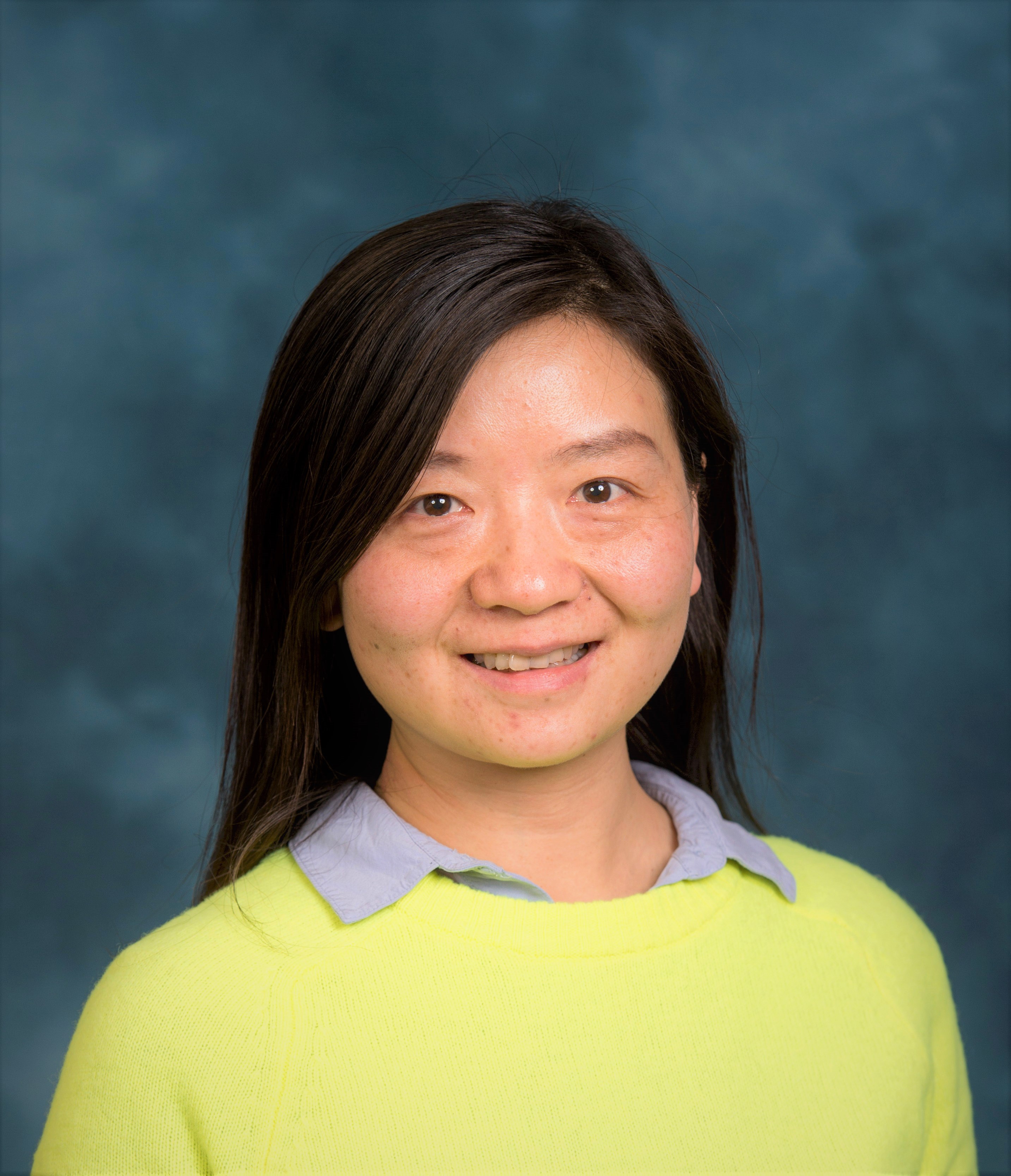}}]{Shan Bao} 
is an Associate Professor and an Associate Research Professor at the University of Michigan. She has led and conducted multiple, large, simulator and naturalistic-driving studies for industry and government sponsors. Her areas of expertise include the statistical analysis of crash datasets and naturalistic data, vulnerable road user safety, experimental design, algorithm development to identify driver states and movement, evaluation of driving-safety technologies, measurement of driver performance, driver decision making, and statistical and stochastic modeling techniques. 
\end{IEEEbiography}

\begin{IEEEbiography}[{\includegraphics[width=1in,height=1.25in,clip,keepaspectratio]{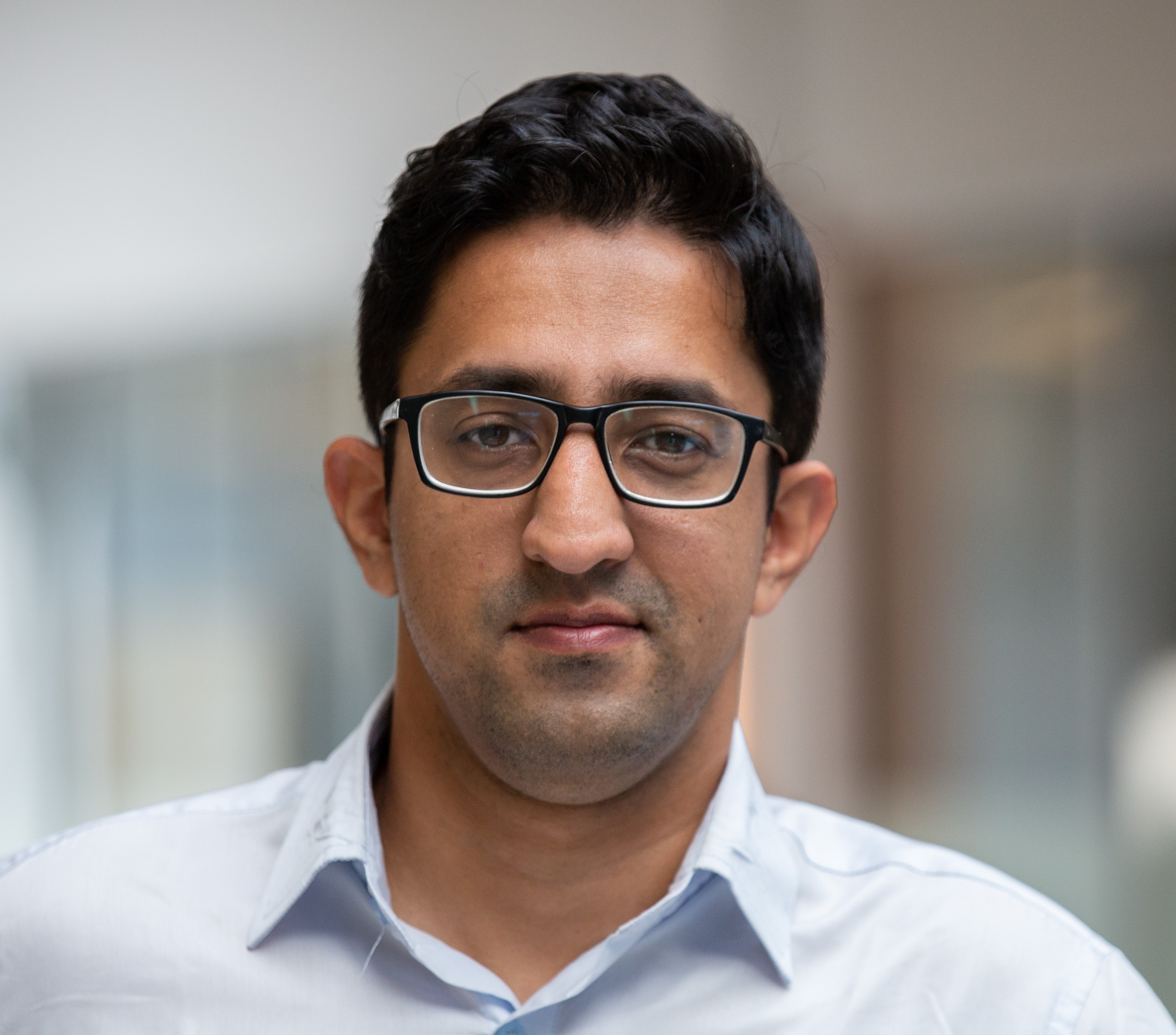}}]{Arpan Kusari}
is an Assistant Research Scientist in the Engineering Systems Group at University of Michigan Transportation Research Institute. He is primarily interested in refining the Automated Driving Systems (ADS) pipeline with a broad impact in terms of safety, efficiency and robustness. He previously spent five years at Ford Motor Company researching exclusively on making autonomous vehicles safe and viable. He received his M.S. in Civil Engineering from Purdue University, USA and his Ph.D. in Civil Engineering from University of Houston, USA.
\end{IEEEbiography}




\end{document}